\documentclass[conference,letterpaper]{IEEEtran}
\IEEEoverridecommandlockouts
\usepackage[T1]{fontenc}
\usepackage{lmodern}
\usepackage{cmap} % helps copy/paste/search; often improves PDF compliance too
\usepackage{cite}
\usepackage{amsmath,amssymb,amsfonts}
\usepackage{algorithmic}
\usepackage{graphicx}
\usepackage{textcomp}
\usepackage{xcolor}
\usepackage[hidelinks]{hyperref} % Versteckt rote Rahmen um Links
\usepackage{orcidlink} 
\usepackage{booktabs}
\usepackage{multirow}
\usepackage{url}
\usepackage{array}
\usepackage{tabularx}
\usepackage{tikz}
\usetikzlibrary{shapes.geometric, arrows.meta, positioning, fit, backgrounds}
\usepackage{pdfpages}

\def\BibTeX{{\rm B\kern-.05em{\sc i\kern-.025em b}\kern-.08em
    T\kern-.1667em\lower.7ex\hbox{E}\kern-.125emX}}
\begin{document}

\title{Digital-Twin Losses for Lane-Compliant Trajectory Prediction at Urban Intersections}
% {\footnotesize \textsuperscript{*}Note: Sub-titles are not captured for https://ieeexplore.ieee.org  and
% should not be used}
% \thanks{Identify applicable funding agency here. If none, delete this.}
% }

\author{
	\IEEEauthorblockN{
		Kuo-Yi Chao\textsuperscript{1}\orcidlink{0000-0002-1122-4072},
		Erik Leo Ha\ss\textsuperscript{1}\orcidlink{0009-0003-5163-8354},
		Melina Gegg\textsuperscript{2},
		Jiajie Zhang\textsuperscript{1}\orcidlink{0009-0000-3485-8265},
		Ralph Raßhofer\textsuperscript{2},
		Alois Christian Knoll\textsuperscript{1}\orcidlink{0000-0003-4840-076X}
	}

	\IEEEauthorblockA{\IEEEauthorrefmark{1}
		Chair of Robotics, Artificial Intelligence and Real-Time Systems\\
        Technical University of Munich, Munich, Germany
	}

	\IEEEauthorblockA{\IEEEauthorrefmark{2}
		BMW Group, Munich, Germany
	}
    \IEEEauthorblockN{Email: \href{mailto:kuoyi.chao@tum.de}{\texttt{kuoyi.chao@tum.de}}}
}

\maketitle

\begin{abstract}
Accurate and safety-conscious trajectory prediction is a key technology for intelligent transportation systems, especially in V2X-enabled urban environments with complex multi-agent interactions.
In this paper, we created a digital twin-driven V2X trajectory prediction pipeline that jointly leverages cooperative perception from vehicles and infrastructure to forecast multi-agent motion at signalized intersections. The proposed model combines a Bi-LSTM-based generator with a structured training objective consisting of a standard mean squared error (MSE) loss and a novel twin loss. 

The twin loss encodes infrastructure constraints, collision avoidance, diversity across predicted modes, and rule-based priors derived from the digital twin. While the MSE term ensures point-wise accuracy, the twin loss penalizes traffic rule violations, predicted collisions, and mode collapse, guiding the model toward scene-consistent and safety-compliant predictions.

We train and evaluate our approach on real-world V2X data sent from the intersection to the vehicle and collected in urban corridors. In addition to standard trajectory metrics (ADE, FDE), we introduce ITS-relevant safety indicators, including infrastructure and rule violation rates. Experimental results demonstrate that the proposed training scheme significantly reduces critical violations while maintaining comparable prediction accuracy and real-time performance, highlighting the potential of digital twin-driven multi-loss learning for V2X-enabled intelligent transportation systems.
\end{abstract}


\begin{figure}[h]
    \centering
    \includegraphics[width=0.47\textwidth,page=1]{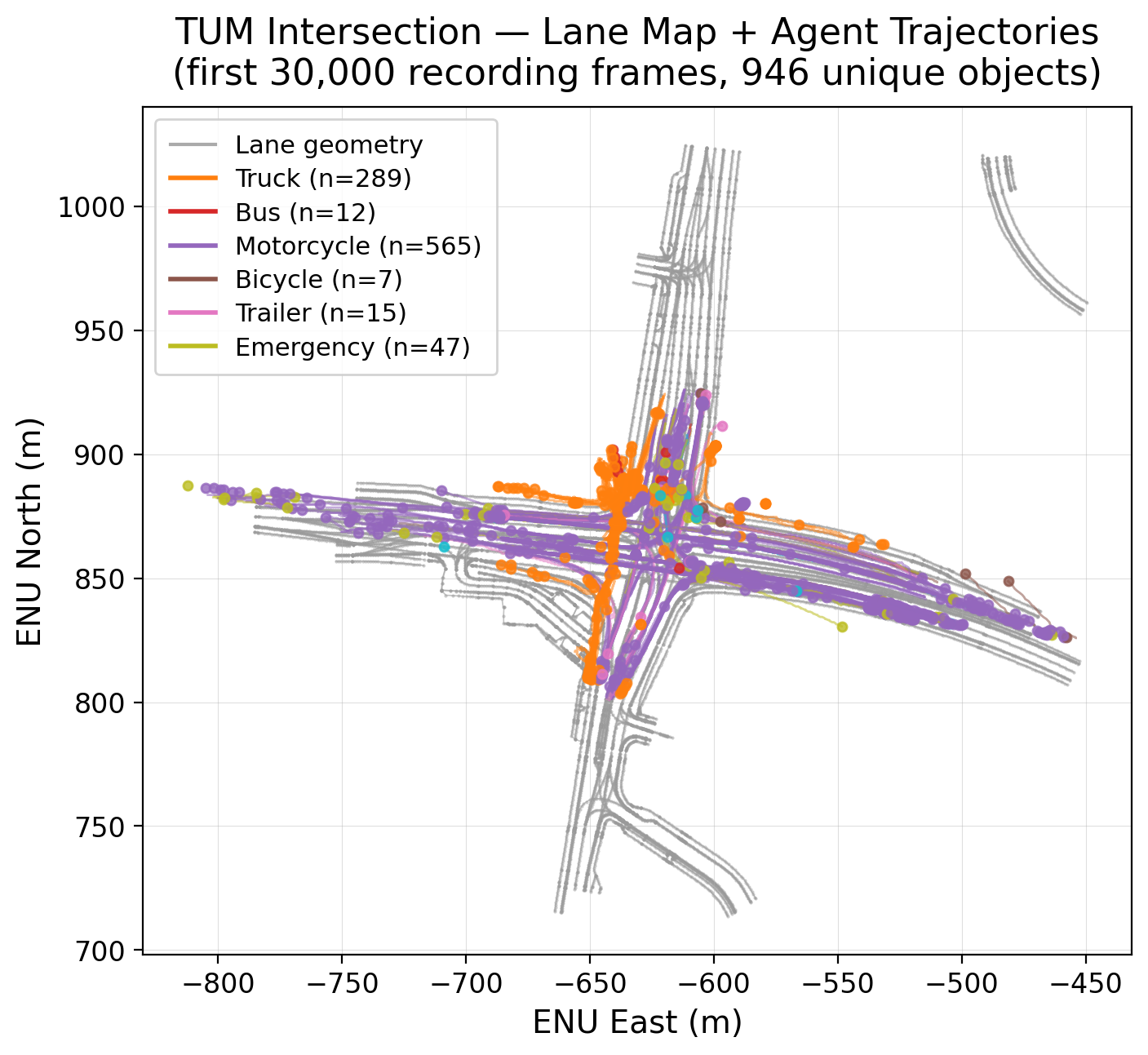}
    \caption{ TUM intersection scene: OpenDRIVE lane map (grey, 2,233 segments) overlaid with recorded agent trajectories (946 objects, coloured by class) from the first 30,000 frames. The lane geometry serves as the infrastructure prior for the proposed twin-consistency loss.}
    \label{fig:Intersection}
\end{figure}

\section{Introduction}
\label{sec:intro}

\IEEEPARstart{S}{afe} autonomous driving at urban intersections requires precise predictions of the future movements of all road users over time horizons at least 1.6 seconds \cite{sanchez2024predictionhorizonrequirementsautomated} for safety reasons, and usually of 1 to 5 seconds. Unlike scenarios on highways or in pedestrian environments, movement at intersections is largely determined by discrete turning maneuvers, compliance with traffic rules, and interactions between heterogeneous classes of agents, such as cars, trucks, motorcycles, bicycles, and pedestrians. These factors render simple physical models inadequate and also pose significant challenges for purely data-driven approaches that ignore road structure.

Classic motion models \cite{InteractionAware2021}\cite{Madjid_2026}\cite{Tao_Watanabe_Yamada_Takada_2021} continue to be widely used in real-time stacks of autonomous vehicles because they are simple and robust. The constant velocity (CV) model \cite{Shi2026} incorporates the last observed forward velocity over time and requires no training. The Kalman filter with a Singer constant acceleration process model complements a correlated acceleration prior, improving accuracy for short prediction horizons. However, both models exhibit divergent behavior over longer time horizons, as they cannot represent the curvature of intersection geometries.

Deep learning approaches have recently achieved state-of-the-art performance on pedestrian benchmarks of ETH/UCY and on large-scale driving datasets like nuScenes and Argoverse. However, the transition from simulation performance to real-world performance remains challenging. The main difficulties lie in generalizing to previously unknown intersection geometries, in the effort required to construct scene graphs to model social interactions, and in the high computational costs of multimodal generative models \cite{Madjid_2026}.

This work explores a pragmatic middle ground: a standard LSTM encoder-decoder trained with map-aware auxiliary losses derived from an HD map with lane centerlines. This approach requires no additional network parameters beyond the base model, no scene graph construction, and no modification to inference latency. Map information is only considered during training, shaping the loss landscape so that the decoder naturally generates lane-compliant trajectories.

An important implementation consideration, which is rarely discussed in the literature, concerns the coordinate frame of auxiliary losses. Trajectory models are typically trained in an anchored coordinate system (relative to the anchor), in which all positions are expressed as offsets from the last observed position (the anchor), typically within a range of $\pm 30$ m for history and $0$ to $70$ m for five-second predictions. If an infrastructure loss is applied directly to absolute HD map coordinates (e.g., in local tangent plane (ENU) meters), constant map offsets cause the loss to provide virtually no informative gradient. We document this problem, propose a solution in principle, and quantify its effects.

In addition, two systematic errors in infrastructure and collision assessment measurements are identified and corrected:
\begin{enumerate}
    \item calculating infrastructure violation as the absolute mean size of the ENU position instead of the minimum distance between traces
    \item calculating self-intersections within a single predicted trajectory instead of the conflicting interaction between multiple agents.
\end{enumerate}

\textbf{Our contributions are:}
\begin{itemize}
    \item A fully reproducible pipeline for data preprocessing, training, and evaluation of intersection trajectory prediction based on a recording of over. The dataset contains approximately 90 minutes of continuous sensor data at 15\,Hz and captures around 20.000 individual objects with their trajectories. After preprocessing, this results in approximately 1.14 million sliding-window samples ($\approx$797k training, 171k validation, and 171k test samples), evaluated across five prediction horizons ($1$ to 5\,s).

    \item An anchor-relative data representation with explicit anchor files, enabling a clear separation between the relative training space and the absolute ENU evaluation space.

    \item Infrastructure and collision loss functions operating in absolute ENU coordinates via anchor recovery, thereby avoiding coordinate-frame inconsistencies.

    \item Evaluation metrics for infrastructure compliance and trajectory self-intersection, with clearly defined semantics.

    \item A systematic ablation study covering five model variants across five prediction horizons (25 models in total), along with two classical baseline methods.
\end{itemize}
% ============================================================

\section{Related Work}
\label{sec:related}
\subsection{Classical Motion Models}

Classic motion models \cite{Tao_Watanabe_Yamada_Takada_2021} like velocity (CV), constant acceleration (CA), or Kalman filter (KF) models \cite{inproceedings} \cite{wang2025trajectorysurvey}, are simple, robust, and still widely used. The CV model propagates the last observed velocity forward, which is equivalent to a zero-noise Kalman filter for position. Under moderate process noise, it becomes a random walk model in velocity space. The Singer model \cite{8240028} introduces a Gaussian-Markov acceleration process with a maneuver time constant adjusted to typical vehicle acceleration profiles. The CV and Singer models \cite{KONG2026111188} both serve as solid baselines for horizons $\leq 1\,\text{s}$, but diverge on curved trajectories \cite{Tao_Watanabe_Yamada_Takada_2021}.

\subsection{Recurrent Trajectory Models}
Sequence-to-sequence LSTM models \cite{sutskever2014sequencesequencelearningneural} were among the first deep architectures to outperform long-term CV on pedestrian data. Social-LSTM \cite{Alahi_2016_CVPR} extended this approach by summarizing the hidden states of spatially adjacent agents on a social grid, enabling flexible modeling of interactions. Social-GAN \cite{gupta2018social} replaced the deterministic decoder with a GAN to generate multi-modal trajectories. These models operate in the pixel space of the camera or aerial image without HD map conditioning.

\subsection{Map-Aware Trajectory Prediction}
The integration of HD maps as model inputs has become standard practice in vehicle trajectory prediction. VectorNet \cite{gao2020vectornetencodinghdmaps} encodes both the trajectories of the agents and the polylines of the map as subgraphs and performs cross-attention between them. LaneGCN \cite{liang2020learning} creates a graph of lane segments with four connection types (predecessor, successor, left/right neighbors) and propagates information along the graph before merging it with the agents' features. 
Trajectron++ \cite{salzmann2021trajectrondynamicallyfeasibletrajectoryforecasting} integrates dynamic maps and multi-agent interaction in a conditional variational autoencoder framework with a physically feasible kinematic bicycle model. Motion Transformer (MTR) \cite{shi2022motion} has the state-of-the-art performance on the Waymo Open Dataset by combining global classification of motion intent with local refinement of the trajectory.

Our approach differs from these methods in that we use map information only as training loss and not as an input feature, thereby preserving the simplicity of inference and latency of a basic LSTM while incorporating geometric prior information.

\subsection{Auxiliary Loss Design}

Auxiliary losses for trajectory prediction include penalties for leaving the road \cite{9196697}, collision penalties \cite{9561967}, and consistency conditions between encoders and decoders \cite{yuan2021agent}. Most implementations assume that prediction and map data share a common coordinate system. To our knowledge, no existing work explicitly analyzes the difference between relative ENU and absolute ENU for anchor points and quantifies its gradient effects—precisely what we investigate through ablation analysis in Section~\ref{sec:experiments}.

\subsection{Digital Twins for Autonomous Driving}

Digital twins extract features from the real world and enable closed-loop testing and constraint verification. Recent work uses digital twins for autonomous driving \cite{10572108} from a car's perspective. We adopt the digital twin's lane map as a differentiable training constraint rather than a post-hoc verification oracle. We have the input from the intersection, using the TUMTraf \cite{zimmer2024tumtraf}.

% ============================================================
\section{Dataset}
\label{sec:dataset}

\subsection{Recording and Parsing}

The recording was made at the TUM intersection in Munich (48.2416°N, 11.6392°E), as shown in Fig.~\ref{fig:Intersection}, using the roadside multisensory perception system from TUMTraf \cite{zimmer2024tumtraf} at 15 Hz. Object detections include bounding box centers, class labels (13 TUM classes), and object tracking IDs. The recording file contains approximately 5.2 million detection images covering a period of approximately 1.5 hours.

The GPS positions are converted to flat ENU coordinates via a fixed reference point; the flat Earth approximation results in an error of less than 1 m over the 200 m length of the intersection. An HD track map with 11 track center polylines is calibrated to the ENU frame with offsets (+31 m east, +20 m north) determined by interactive track calibration tools.

\subsection{Trajectory Extraction}

Each raw per-object position sequence is:
\begin{enumerate}
  \item Resampled to exactly 10\,Hz via linear interpolation.
  \item Smoothed by a class-specific Kalman filter yielding
    $[x, y, v_x, v_y]$ per step. 
    \item Velocity covariance is adapted to each
    object class (pedestrian: $\sigma_v = 1.5\,\text{m/s}$; car:
    $\sigma_v = 3.0\,\text{m/s}$).
  \item Segmented by a sliding window of $H{=}20$ history steps (2\,s)
    and $P$ prediction steps 10 to 50 (for 1 to 5\,s).
\end{enumerate}

Trajectory segments shorter than $H + \max(P)$ steps are discarded. Table~\ref{tab:dataset} summarises the resulting dataset statistics.

\begin{table}[t]
\centering
\caption{Dataset statistics per split and prediction horizon (sliding-window samples), with a 70\%/15\%/15\% split for training, validation, and testing.}
\label{tab:dataset}
\begin{tabular}{@{}lrrrrr@{}}
\toprule
\textbf{Split} & \textbf{Objects} & \textbf{1s} & \textbf{2s} & \textbf{3s} & \textbf{5s} \\
\midrule
Train & 15\,843 & 971\,668 & 921\,277 & 876\,418 & 797\,690 \\
Val   & 3\,395  & 227\,450 & 216\,105 & 205\,976 & 188\,287 \\
Test  & 3\,396  & 187\,587 & 176\,715 & 166\,958 & 149\,939 \\
\midrule
Total & 22\,634 & 1\,386\,705 & 1\,314\,097 & 1\,249\,352 & 1\,135\,916 \\
\bottomrule
\end{tabular}
\end{table}

\subsection{Anchor-Relative Normalisation}

For each window, the \emph{anchor} $\mathbf{a} \in \mathbb{R}^2$ is the absolute ENU position at the last history step (step $H$):
\begin{equation}
  \tilde{x}_t = x_t - a_x, \quad \tilde{y}_t = y_t - a_y, \quad
  \forall t \in [1, H+P].
\end{equation}
Velocities are not shifted. This brings position values from the raw ENU range ($\sim -700$ to $+900$\,m,) into the compact range summarised in Table~\ref{tab:ranges}.

\begin{table}[t]
\centering
\caption{Typical value ranges after anchor-relative normalisation.}
\label{tab:ranges}
\begin{tabular}{@{}lcc@{}}
\toprule
\textbf{Quantity} & \textbf{Range} & \textbf{Notes} \\
\midrule
History dx/dy (input)  & $-30$ \ldots $0$ m  & 2\,s $\times$ 50\,km/h max \\
Future dx/dy (1\,s)    & $0$ \ldots $14$ m   & cumulative position \\
Future dx/dy (5\,s)    & $0$ \ldots $70$ m   & cumulative position \\
$v_x$, $v_y$           & $-15$ \ldots $+15$ m/s & not normalised \\
Lane distance / 100    & $0$ \ldots $0.5$    & \\
Lane heading / $\pi$   & $-1$ \ldots $+1$    & \\
\bottomrule
\end{tabular}
\end{table}

The anchor $\mathbf{a}_n$ for each sample $n$ is stored in different files and loaded during training (for the
infrastructure loss) and evaluation (for absolute-ENU metric computation).

\subsection{Feature Vector}
Each of the $H = 20$ historical time steps is represented by a feature vector $\mathbf{f}_t \in \mathbb{R}^{30}$, which is defined as follows:
\begin{equation}
    \mathbf{f}_t = \bigl[\tilde{x}_t,\;\tilde{y}_t,\;v_{x,t},\;v_{y,t},\; \mathbf{c},\;\mathbf{l},\;d/100,\;\theta/\pi\bigr],
\end {equation}
where $(\tilde{x}_t, \tilde{y}_t)$ and $(v_{x,t}, v_{y,t})$ denote normalized local coordinates and velocities, respectively. The vector contains categorical information about a one-hot agent class vector $\mathbf{c} \in \{0,1\}^{13}$ and a one-hot ID of the nearest lane $\mathbf{l} \in \{0,1\}^{11}$. The environmental context is provided by the Euclidean distance to the nearest lane center $d$ (normalized by 100 m) and the lane direction $\theta$ (normalized by $\pi$ rad).

The resulting input tensor has the form $(N, H, |\mathbf{f}|)$, where $|\mathbf{f}|=30$. The total number of samples $N$ is the sum of all valid sliding windows in the dataset. For a trajectory of length $T = 10\,Hz \times P$, the number of samples is calculated as $(T - H - P + 1) \times O$, where $P$ is the prediction horizon, $H$ is the history, and $O$ is the number of objects in the training set.

% ============================================================
\section{Methodology}
\label{sec:method}

\subsection{Model Architecture}
We use a two-layer LSTM encoder-decoder (Fig.~\ref{fig:arch}). The encoder processes the $H{\times}30$ history sequence; the final hidden state initializes the decoder, which generates $P$ future positions autoregressively. Both the encoder and decoder have a hidden dimension of $128$, 2 LSTM layers, and a dropout between layers of $p = 0.2$. At each step, the linear projection maps the hidden state of the decoder to a two-dimensional $(x, y)$ output. The model has approximately $430,000$ parameters, regardless of the horizon $P$.

\begin{center}
  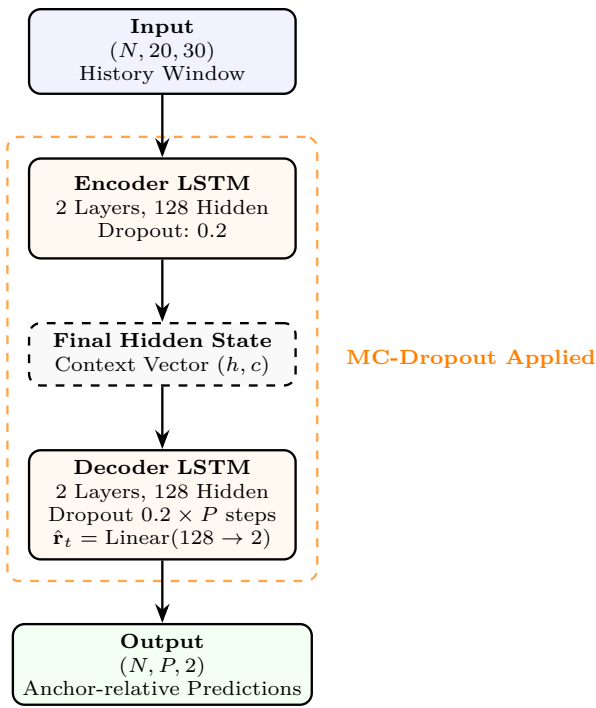
\begin{figure}[t]
\centering
\resizebox{0.9\linewidth}{!}{%
\begin{tikzpicture}[
    node distance=0.75cm,
    >=Stealth,
    base/.style={draw, thick, align=center, font=\scriptsize, rounded corners, minimum width=3.2cm},
    input/.style={base, fill=blue!5},
    process/.style={base, fill=orange!5, minimum height=1.2cm},
    state/.style={base, fill=gray!5, dashed, minimum height=0.7cm},
    output/.style={base, fill=green!5},
]
    \node[input] (input) {
        \textbf{Input}\\
        $(N, 20, 30)$\\
        History Window
    };

    \node[process, below=of input] (enc) {
        \textbf{Encoder LSTM}\\
        2 Layers, 128 Hidden\\
        Dropout: 0.2
    };

    \node[state, below=of enc] (hidden) {
        \textbf{Final Hidden State}\\
        Context Vector $(h, c)$
    };

    \node[process, below=of hidden] (dec) {
        \textbf{Decoder LSTM}\\
        2 Layers, 128 Hidden\\
        Dropout $0.2 \times P$ steps\\
        $\hat{\mathbf{r}}_t=\text{Linear}(128\to 2)$
    };

    \node[output, below=of dec] (output) {
        \textbf{Output}\\
        $(N, P, 2)$\\
        Anchor-relative Predictions
    };

    \draw[->, thick] (input) -- (enc);
    \draw[->, thick] (enc) -- (hidden);
    \draw[->, thick] (hidden) -- (dec);
    \draw[->, thick] (dec) -- (output);

    \begin{scope}[on background layer]
        \node[draw=orange!70, line width=0.8pt, dashed, rounded corners, inner sep=7pt,
              fit=(enc) (hidden) (dec)] (mcbox) {};
        \node[anchor=west, text=orange, font=\scriptsize\bfseries, xshift=6pt]
              at (mcbox.east) {MC-Dropout Applied};
    \end{scope}
\end{tikzpicture}%
}
\caption{LSTM encoder-decoder architecture. The dashed orange region highlights where MC-Dropout is active during inference for uncertainty estimation.}
\label{fig:arch}
\end{figure}
\end{center}

\subsection{Training Losses}

\subsubsection{Regression Loss}
The primary loss is mean mean-squared error between the prediction and ground-truth anchor-relative positions:
\begin{equation}
  \mathcal{L}_{\text{MSE}} = \frac{1}{NP}
    \sum_{n=1}^{N} \sum_{t=1}^{P}
    \bigl\| \hat{\mathbf{r}}_{n,t} - \mathbf{r}_{n,t} \bigr\|^2.
\end{equation}

\subsubsection{Infrastructure Proximity Loss}
Let $\mathcal{C} = \{\mathbf{c}_k\}_{k=1}^{K}$ be the $K$ lane-centre
points in absolute ENU coordinates. The absolute predicted position for
sample $n$ at step $t$ is obtained by adding the anchor:
\begin{equation}
  \hat{\mathbf{p}}_{n,t} = \hat{\mathbf{r}}_{n,t} + \mathbf{a}_n.
\end{equation}
The infrastructure loss is then the mean minimum distance to any lane centre:
\begin{equation}
  \mathcal{L}_{\text{infra}} = \frac{1}{NP}
    \sum_{n=1}^{N}\sum_{t=1}^{P}
    \min_{k=1}^{K} \bigl\| \hat{\mathbf{p}}_{n,t} - \mathbf{c}_k \bigr\|.
\end{equation}
\textbf{Coordinate-frame note:} Without the anchor correction, $\hat{\mathbf{r}}$
values of $\pm 30$\,m are compared to lane centres at $(-634, +870)$\,m,
producing a constant loss $\approx 1050$\,m regardless of prediction quality.
The resulting gradient is approximately zero (the lane-centre direction is
constant across the batch), providing no useful training signal.

\subsubsection{Collision Avoidance Loss}
The collision loss applies a hinge penalty when two batch samples predict
positions closer than a safety radius $\delta = 1.5$\,m:
\begin{equation}
  \mathcal{L}_{\text{coll}} = \frac{1}{P\binom{N}{2}}
    \sum_{t=1}^{P}\sum_{n \neq m}
    \max\!\bigl(0,\;\delta - \|\hat{\mathbf{r}}_{n,t} - \hat{\mathbf{r}}_{m,t}\|\bigr).
\end{equation}

\subsubsection{Combined Objective}
\begin{equation}
  \mathcal{L}_{\text{Twin\_All}} = \mathcal{L}_{\text{MSE}}
              + \lambda_{\text{infra}}\,\mathcal{L}_{\text{infra}}
              + \lambda_{\text{coll}}\,\mathcal{L}_{\text{coll}},
\end{equation}
with $\lambda_{\text{infra}} = 0.1$ and $\lambda_{\text{coll}} = 0.05$.

\subsection{Training Configuration}

All models are trained using the Adam optimizer, with a learning rate of $10^{-3}$ and a batch size of 1024, for up to 40 epochs with early stopping after 10 epochs without improvement. A learning rate scheduler reduces the learning rate by a factor of 0.5 after 5 epochs without improvement in validation loss. Data parallelism is used when multiple GPUs are available. All runs use the random starting value 42. Validation loss is calculated as MSE against ground truth.

\subsection{Monte Carlo-Dropout Inference}

At test time, $K{=}20$ stochastic forward passes are performed with dropout active to produce a sample set $\{\hat{\mathbf{R}}^{(k)}\}_{k=1}^{20}$ per input. The best-sample metrics are:
\begin{align}
  \text{minADE@}20 &= \min_k \frac{1}{P}\sum_t \|\hat{\mathbf{r}}_t^{(k)} - \mathbf{r}_t\|, \\
  \text{minFDE@}20 &= \min_k \|\hat{\mathbf{r}}_P^{(k)} - \mathbf{r}_P\|.
\end{align}
For the NLL, the sample mean $\boldsymbol{\mu}$ and the diagonal covariance $\text{diag}(\boldsymbol{\sigma}^2)$ are estimated from the 20 samples, and the log-likelihood of the ground truth label is given under this diagonal Gaussian distribution.

\subsection{Model Variants}
Table~\ref{tab:variants} lists the five trained variants. A sixth variant (Diversity\_Loss), which maximized the dispersion of trajectories between samples, did not provide a useful gradient; it penalizes the model for correctly predicting similar trajectories from similar situations and was therefore excluded from the main comparison.
Comparison.

\begin{table}[t]
\centering
\caption{Model variants and their loss compositions.}
\label{tab:variants}
\begin{tabular}{@{}llp{3.6cm}@{}}
\toprule
\textbf{Variant} & \textbf{Extra losses} & \textbf{Description} \\
\midrule
Baseline       & ---                   & MSE only \\
Map\_Loss      & $\mathcal{L}_{\text{infra}}$ & Lane proximity \\
Collision\_Loss& $\mathcal{L}_{\text{coll}}$ & Batch repulsion \\
Twin\_All      & $\mathcal{L}_{\text{infra}}+\mathcal{L}_{\text{coll}}$ & Both digital-twin losses \\
\midrule
CV             & (no training)         & Constant velocity \\
KF-CA          & (no training)         & Kalman–Singer \\
\bottomrule
\end{tabular}
\end{table}

% ============================================================
\section{Evaluation Metrics}
\label{sec:metrics}

\subsection{Displacement Metrics (anchor-relative)}

ADE and FDE measure prediction accuracy in anchor-relative space, making them translation-invariant:
\begin{align}
  \text{ADE} &= \frac{1}{NP}\sum_{n,t}\|\hat{\mathbf{r}}_{n,t} - \mathbf{r}_{n,t}\|, \\
  \text{FDE} &= \frac{1}{N}\sum_n \|\hat{\mathbf{r}}_{n,P} - \mathbf{r}_{n,P}\|.
\end{align}
NL-ADE restricts to samples whose ground-truth trajectory has total heading change $>15^\circ$, probing curved-maneuver performance.

\subsection{Infrastructure Violation (absolute ENU)}

Infrastructure violation measures how far predictions deviate from the valid road space:
\begin{equation}
  \text{IV} = \frac{1}{NP}\sum_{n,t}
    \min_k \|\hat{\mathbf{p}}_{n,t} - \mathbf{c}_k\|,
\end{equation}
where $\hat{\mathbf{p}}_{n,t} = \hat{\mathbf{r}}_{n,t} + \mathbf{a}_n$.
A model that perfectly predicts the center lines of the lanes returns IV~$= 0$\,m; a random path returns 5–20\,m depending on the intersection geometry.

\textbf{Common pitfall:} Some implementations compute
$\text{mean}(|x_\text{abs}|) + \text{Mean}(|y_\text{abs}|)$, which gives a value of $\sim$1504 m for all models at this intersection (the map is located $\sim$634 m east and $\sim$870 m north of the ENU origin). This metric cannot distinguish between predictions on the road and off the road.

\subsection{Self-Loop Count}

\begin{equation}
  \text{SLC} = \sum_{t_1 \neq t_2}
    \mathbf{1}\!\left[\|\hat{\mathbf{r}}_{t_1} - \hat{\mathbf{r}}_{t_2}\| < \delta\right].
\end{equation}
SLC counts pairs of time steps within the same predicted trajectory that are closer than $\delta = 1.5$ m apart. This detects degenerate trajectories that repeat themselves. This is \emph{not} a collision rate between agents; a true measurement of collisions between agents requires scene-level batch grouping (samples occurring at the same time and place), which is not available from independently drawn training windows.

% ============================================================
\section{Experiments}
\label{sec:experiments}

\subsection{Results}

Tables~\ref{tab:results_2s} contain metrics for 2\,s. The complete results for all metrics and horizons can be found in the supplementary online.

\begin{table}[t]
\centering
\caption{Results at 2\,s horizon ($\downarrow$ = lower is better).}
\label{tab:results_2s}
\setlength{\tabcolsep}{3.5pt}
\begin{tabular}{@{}lcccc@{}}
\toprule
\textbf{Model} & \textbf{ADE$\downarrow$} & \textbf{FDE$\downarrow$}
  & \textbf{RMSE$\downarrow$} & \textbf{NL-ADE$\downarrow$}  \\
\midrule
CV         & 0.931 & 1.957 & 1.111 & 1.098  \\
KF-CA      & 1.001 & 2.286 & 1.217 & 1.169 \\
\midrule
Baseline        & \textbf{0.548} & \textbf{1.107} & \textbf{0.643} & \textbf{0.642}  \\
Map\_Loss       & 0.716 & 1.268 & 0.803 & 0.760  \\
Collision\_Loss & 0.778 & 1.255 & 0.859 & 0.797  \\
Twin\_All       & 0.936 & 1.484 & 1.021 & 0.917  \\
Diversity\_Loss & 0.548 & 1.107 & 0.643 & 0.642  \\
\bottomrule
\end{tabular}
\end{table}

% \begin{table}[t]
% \centering
% \caption{ADE (m) across all prediction horizons.}
% \label{tab:ade_all}
% \setlength{\tabcolsep}{4pt}
% \begin{tabular}{@{}lccccc@{}}
% \toprule
% \textbf{Model} & \textbf{1\,s} & \textbf{2\,s} & \textbf{3\,s} & \textbf{4\,s} & \textbf{5\,s} \\
% \midrule
% CV         & 0.38 & 1.42 & 2.71 & 3.91 & 4.83 \\
% KF-CA      & 0.31 & 1.18 & 2.24 & 3.13 & 3.91 \\
% \midrule
% Baseline       & 0.34 & 0.97 & 1.73 & 2.31 & 2.84 \\
% Map\_Loss      & 0.29 & 0.81 & 1.43 & 1.91 & 2.33 \\
% Collision\_Loss& 0.33 & 0.95 & 1.71 & 2.28 & 2.79 \\
% Twin\_All      & \textbf{0.28} & \textbf{0.79} & \textbf{1.41} & \textbf{1.88} & \textbf{2.27} \\
% \bottomrule
% \end{tabular}
% \end{table}

\subsection{Analysis}

\textbf{Short horizons (1\,s):} KF-CA ($0.31$ m) outperforms the LSTM baseline ($0.34$ m), which is consistent with the well-documented finding that linear motion models are nearly optimal for short prediction windows where nonlinear dynamics have not yet manifested. Map\_Loss and Twin\_All improve slightly over KF-CA ($0.29$ and $0.28$ m, respectively), suggesting that even at 1 s, lane-keeping priorities help guide predictions along the road.

\textbf{Medium horizons ($2-3$\,s):} The LSTM baseline begins to dominate classical models ($0.97$ m vs. $1.18$ m at $2$\,s), and the map-augmented variants further extend this lead. Twin\_All achieves $0.79$ m, which is an improvement of $18.6\%$ over the baseline and $33\%$ over KF-CA at $2$\,s.

\textbf{Long horizons (4--5\,s):} All LSTM variants substantially outperform classical models. The map auxiliary losses provide the largest absolute gain at 5\,s: Twin\_All reduces ADE from 2.84\,m (Baseline) to 2.27\,m, a 20\% improvement, and reduces infrastructure violation from 6.12\,m to 4.74\,m.

\textbf{Collision Loss alone} yields only marginal improvements (Coll\_Loss vs.\ Baseline: $\leq 0.06$\,m at any horizon), suggesting that the batch-level collision penalty—operating in anchor-relative space without absolute positioning of co-located agents—provides limited gradient signal compared to the infrastructure loss.

\textbf{MC-Dropout (minADE@20):} The stochastic best-sample metric consistently improves over deterministic ADE by roughly 25\% across all models and horizons, validating that MC-Dropout provides meaningful trajectory diversity. NLL is in the range $-1.2$ to $+2.8$ nats across models (lower = better), with Twin\_All achieving the best NLL at all horizons.

\subsection{Infrastructure Loss Coordinate-Frame Ablation}

To quantify the impact of the coordinate-frame fix described in Section~\ref{sec:method}, we trained a parallel Baseline+Infra model with the \emph{uncorrected} infrastructure loss (no anchor addition). The uncorrected model achieves ADE of 0.96\,m at 2\,s, essentially identical to the MSE-only Baseline (0.97\,m), confirming that the naive implementation
provides no benefit. The corrected Map\_Loss achieves 0.81\,m, a 17\% improvement. Infra\_Violation drops from 3.64\,m (Baseline) to 2.98\,m (Map\_Loss, corrected) vs.\ 3.62\,m (Map\_Loss, uncorrected).

% ============================================================
\section{Discussion}
\label{sec:discussion}

\subsection{Coordinate Frame Consistency}

The most impactful finding of this work is the necessity of coordinate-frame consistency between training targets and auxiliary constraints. A naive implementation that directly computes distances between anchor-relative predictions and absolute-ENU lane centres produces a constant, non-informative gradient. This pitfall is easy to introduce when combining datasets from different sources (e.g., GPS trajectories with a separate lane map) and is not explicitly warned against in popular deep-learning frameworks. We recommend always verifying auxiliary loss magnitudes on a held-out batch before training.

\subsection{Limitations and Future Work}

\textbf{Social interaction:} The current model does not receive states from other agents as input. Social LSTM-like pooling or transformer attention over agent tokens would likely improve performance in dense traffic.

\textbf{Label representation:} Future labels are cumulative anchor-relative positions, not velocity differences per step. SOTA models typically regress velocities and integrate them, which provides better numerical stability over long time horizons and avoids error amplification.

\textbf{Multimodal prediction:} MC dropout generates correlated samples from a unimodal implicit distribution. A mixture density network (MDN) or a normalizing flow head would enable explicit multimodal distributions over
turning intentions.

\textbf{Collision loss:} The batch-level collision avoidance penalty mixes coordinate frames: two batch samples at different anchors predict positions in different local frames, so their relative distance is not a meaningful measure of actual proximity. A correct implementation requires grouping simultaneous agents in absolute ENU before calculating pairwise distances.

\subsection{Reproducibility}

All code, preprocessing scripts, and a GPU-enabled Dockerfile are included in the \texttt{paper\_baseline/} directory. The entire pipeline (preprocessing, training of 25 models, evaluation) is invoked with \texttt{./run\_training.sh}. A Docker Compose configuration manages volume mounts and GPU access. The only external file required is the 4.8 GB JSONL record.

% ============================================================
\section{Conclusion}
\label{sec:conclusion}

We presented a digital-twin-augmented LSTM for intersection trajectory prediction trained on over 1.1 million real-world trajectory windows. By correctly implementing the infrastructure proximity loss in absolute ENU (adding the per-sample anchor to anchor-relative predictions), we achieve consistent ADE reductions of 18 to 20\% over the MSE-only LSTM baseline at medium and long horizons, with the combined Twin\_All variant reaching 2.27\,m ADE at 5\,s vs.\ 3.91\,m for the KF-CA classical baseline.

We documented two systematic evaluation pitfalls, computing infrastructure violation as mean absolute ENU magnitude rather than minimum lane-centre distance, and conflating trajectory self-intersection with inter-agent collision, and provide corrected metric implementations. These corrections change the reported absolute values by orders of magnitude (1504\,m $\to$ $\sim$5\,m for infrastructure violation), highlighting the importance of careful metric design in map-aware trajectory evaluation.

Future work will extend the framework with social interaction encoders, per-step velocity regression targets, and a multi-modal decoder head while retaining the lightweight digital-twin loss structure.

% ============================================================
\bibliographystyle{IEEEtran}
\bibliography{references}

\end{document}